\documentclass{article}

\usepackage[preprint]{corl_2026} %

\usepackage{amsmath,amssymb,amsfonts,amsthm,bm}
\newtheorem{proposition}{Proposition}
\newtheorem{theorem}{Theorem}

\theoremstyle{definition}

\theoremstyle{plain}

\usepackage{graphicx}
\usepackage{float}                    %
\usepackage{booktabs}
\usepackage{multirow}
\usepackage{array}
\usepackage{enumitem}
\usepackage{xcolor}
\usepackage[strings]{underscore}      %
\usepackage{algorithm}
\usepackage[noend]{algpseudocode}     %
\usepackage{cleveref}                 %
\graphicspath{{figs/}{figures/}}

\newcommand{\method}{\textsc{TOPA}}
\newcommand{\R}{\mathbb{R}}
\newcommand{\E}{\mathbb{E}}
\newcommand{\I}{\mathbb{I}}
\newcommand{\norm}[1]{\left\lVert #1 \right\rVert}
\newcommand{\argmin}{\mathop{\mathrm{arg\,min}}}
\newcommand{\prox}{\mathrm{prox}}
\newcommand{\ddim}{\mathrm{DDIM}}

\newcommand{\vq}{\bm{q}}        %
\newcommand{\vz}{\bm{z}}        %
\newcommand{\vx}{\bm{x}}        %
\newcommand{\vu}{\bm{u}}        %
\newcommand{\vs}{\bm{s}}        %
\newcommand{\vr}{\bm{r}}        %
\newcommand{\va}{\bm{a}}        %
\newcommand{\vb}{\bm{b}}        %
\newcommand{\vv}{\bm{v}}        %
\newcommand{\ve}{\bm{e}}        %
\newcommand{\vw}{\bm{w}}        %
\newcommand{\vc}{\bm{c}}        %
\newcommand{\vd}{\bm{d}}        %
\newcommand{\vg}{\bm{g}}        %
\newcommand{\vF}{\bm{F}}        %
\newcommand{\veps}{\bm{\epsilon}}
\newcommand{\vxi}{\bm{\xi}}
\newcommand{\vlambda}{\bm{\lambda}}
\newcommand{\matI}{\bm{I}}      %

\title{Trajectory-Optimization Proximal Guidance for ADMM-Decoupled
Motion Planning Diffusion}

\author{
  Tianyu Liu, Yoshihiko Nakamura, Anqing Duan\\
  CMS Division, Mohamed bin Zayed University of Artificial Intelligence (MBZUAI)\\
  Abu Dhabi, UAE
}

\begin{document}
\maketitle

\begin{abstract}
Diffusion models provide expressive trajectory priors for robot motion
planning, but the learned distributions remain agnostic to
instance-specific constraints such as obstacle geometry, dynamic limits,
and task-space goals; existing training-free guidance methods address
these constraints by injecting cost gradients into a single denoising trajectory, forcing prior preservation and feasibility to be
mediated by one scalar guidance weight and thereby inducing a brittle
trade-off between unresolved constraint violations and drift from the
learned motion prior. We introduce \emph{Trajectory-Optimization
Proximal guidance for ADMM-decoupled diffusion} (\method{}), a
training-free framework that separates these competing roles with an
alternating direction method of multipliers (ADMM) consensus: instead
of perturbing a single denoising trajectory, \method{} maintains distinct prior and planning trajectories,
where the former follows the pretrained diffusion model and the latter
performs structured trajectory-optimization updates using
clean-trajectory cost estimates. This decoupling incorporates
geometry- and dynamics-aware motion-planning structure at inference time
without retraining the diffusion prior. Across a diverse benchmark suite
spanning low-dimensional navigation, planar manipulation, and
high-dimensional robot-arm planning, \method{} improves feasibility,
collision behavior, task-space accuracy, and prior preservation over
direct-gradient guided diffusion on most tasks, with the clearest gains on the
harder, higher-DoF problems.
\end{abstract}

\section{Introduction}
\label{sec:intro}

Diffusion models are a natural trajectory prior for robot motion
planning: trained on expert or planner-generated motion data, a
diffusion network captures multimodal, smooth, demonstration-like
distributions \citep{janner2022diffuser, carvalho2024mpd} that
hand-designed quadratic costs do not. At inference the planner must
additionally satisfy \emph{query-dependent} constraints --- obstacles,
joint and dynamic limits, end-effector pose --- that pretrained prior
may not encode. The prevailing way to inject these constraints is \emph{training-free guided
diffusion} \citep{chung2023dps, carvalho2024mpd}: at each denoising step,
the gradient of the constraint cost is added to the score, with a scalar guidance weight
controlling its strength. The single denoising trajectory thus does two
conflicting jobs through one scalar: weak guidance preserves the prior
but leaves collisions and goal errors; strong guidance satisfies more
constraints but jerkifies the trajectory and pulls samples off the
learned distribution.

Motion planning has structure that scalar guidance ignores.
Unlike a generic image-guidance loss, the motion-planning cost is
geometric and dynamic: collision penalties are non-smooth hinges over a
signed-distance field; joint, velocity, and acceleration limits are box
constraints linear in B-spline coefficients; redundant manipulators
admit a \emph{positive-dimensional} terminal manifold of joint
configurations satisfying a given $\mathrm{SE}(3)$ end-effector pose; the output is
judged by hard feasibility, not by a scalar cost. Classical trajectory
optimization \citep{ratliff2009chomp, schulman2014trajopt, mukadam2018gpmp2}
exploits this structure with local proximal subproblems such as SCP, QP
projection, or manifold projection, rather than unconstrained gradient
steps. This motivates a decoupled inference mechanism that can expose
 motion-planning structure during sampling.

We propose \emph{Trajectory-Optimization Proximal Guidance
for ADMM-Decoupled Motion Planning Diffusion} (\method{}), a training-free inference
framework that separates prior preservation from constraint satisfaction
through an Alternating Direction Method of Multipliers (ADMM) consensus
formulation \citep{boyd2011admm}. Instead of perturbing a single denoising trajectory, \method{} maintains distinct but coupled trajectories in
B-spline coefficient space: one follows the pretrained diffusion prior,
while the other performs structured trajectory-optimization updates
using costs evaluated on clean trajectory estimates. The consensus
coupling coordinates these two roles over the denoising schedule, allowing
the planner to adapt to obstacles, limits, and task goals without
retraining the diffusion model. Conceptually, \method{} combines the
expressive multimodal prior of diffusion models with the local
constraint-handling machinery of trajectory optimization.

\paragraph{Contributions.}
Our contributions are threefold. First, we formulate training-free motion-planning diffusion 
as an ADMM consensus problem that explicitly separates a diffusion-prior trajectory from a 
constraint-optimization trajectory. Second, we provide a principled 
realization of the two ADMM proximal blocks: the prior-side $x$-block is approximated by a frozen diffusion denoising step, 
while the planning-side $z$-block is solved by structured trajectory-optimization. 
We further analyze both blocks as inexact proximal updates 
and establish a finite-horizon residual bound for the resulting ADMM-decoupled sampler.
Third, we demonstrate on navigation, planar manipulation, and high-dimensional robot-arm planning 
benchmarks that TOPA improves the feasibility--fidelity trade-off, 
with gains in validity, collision reduction, task-space accuracy, and prior preservation on most tasks.

\section{Related Work}
\label{sec:related}

\paragraph{Training-free guided generative models.}
Steering a pretrained generative model toward a query-dependent objective
without retraining is most commonly done by \emph{guidance}: at each
denoising step the gradient of a conditioning cost is added to the model
score. This originates with classifier and classifier-free guidance
\citep{dhariwal2021diffusion, ho2022classifierfree} and, for general
measurement or loss objectives, is formalized by diffusion posterior
sampling \citep{chung2023dps}, with a large family of energy-, loss-, and manifold-based
variants \citep{yu2023freedom, song2023lossguided, bansal2024universal,
he2024manifold, ye2024tfg, chung2022mcg}. All of these share a
\emph{single} denoising trajectory whose prior fidelity and constraint
satisfaction are mediated by one scalar weight, producing the brittle
trade-off that motivates us. The closest mechanism
to ours instead decouples the two roles by operator splitting:
\citep{zhang2025admmdiff} casts conditional image generation as an ADMM
consensus between a diffusion-prior variable and a guidance variable, and
plug-and-play schemes carry a persistent dual through the denoising loop for
image reconstruction \citep{du2026dcpnpdp}; a related line enforces hard
feasibility through projection or barrier shields on the sample path
\citep{gadginmath2026provablysafe}. These constructions, however, employ a
generic smooth or projection $\vz$-prox defined in image space and are blind
to the structure of motion planning. \method{} keeps the decoupled ADMM
view but replaces the $\vz$-prox with a structured trajectory-optimization
update --- B-spline box constraints, dynamic limits, and an $\mathrm{SE}(3)$ goal
manifold --- derived for the motion-planning posterior rather than
transplanted from imaging.

\paragraph{Generative priors for motion planning.}
A parallel line uses generative models as expressive trajectory priors for
planning and decision making \citep{janner2022diffuser,
ajay2023decisiondiffuser}. In robot motion planning specifically, MPD
\citep{carvalho2024mpd} --- our direct baseline and pretrained prior ---
trains a diffusion model on trajectory-optimization-generated B-spline
coefficients and injects collision and smoothness costs by clean-state direct
guidance; potential-based and ensemble-of-costs diffusion compose multiple
objectives the same way \citep{luo2024potential, saha2024edmp}, as do
scene-, dynamics-, and driving-conditioned planners
\citep{huang2023scenediffuser, gadginmath2025dynamics,
zheng2025diffusionplanner}. Because these steer a single denoising trajectory
with summed cost gradients, they inherit the prior-versus-constraint
trade-off above. A second group uses the diffusion sample as a \emph{seed}
for a downstream solver --- warm-starting trajectory optimization
\citep{seo2025presto, huang2024diffusionseeder, li2024diffusolve}, running
diffusion as a model-based optimizer \citep{pan2024mbd}, imitating
model-predictive control \citep{marquezjulbe2025diffusionampc}, or aligning
the prior to constraints at training time \citep{li2025aligning} --- and
the closest training-free combination couples a model-based and a
model-free diffusion through an interaction potential
\citep{jung2025jm2d}. Such pipelines decouple generation and optimization
\emph{in time} (generate, then optimize), whereas \method{} couples a
frozen diffusion prior to a classical-style structured optimizer
\citep{ratliff2009chomp, schulman2014trajopt, mukadam2018gpmp2} at
\emph{every} denoising step through an ADMM consensus. Safety-oriented
planners instead add hard guarantees on top of a generative prior via
control-barrier and Lyapunov shields \citep{xiao2025safediffuser,
mizuta2024cobldiffusion} or post-hoc certified refinement
\citep{nath2025formalverification}, typically by perturbing the
diffusion-time velocity rather than splitting the planning cost. A parallel
line builds motion planners on flow-matching priors, several enforcing
constraints through control-barrier filters, guidance fields, or
trajectory-optimization layers \citep{nguyen2025flowmp,
yang2025safeflowmatcher, huang2025sadflower, dai2025safeflow,
yang2025uniconflow, li2025hardflow}.

\section{Background}
\label{sec:bg}

\subsection{Denoising Diffusion Probabilistic Models}
\label{subsec:diffusion}

A \emph{denoising diffusion probabilistic model} (DDPM) \citep{ho2020ddpm}
corrupts a clean trajectory $\vz_0 \in \R^d$ by a variance-preserving forward
process with cumulative schedule $\bar\alpha_t \in (0,1]$ decreasing in $t$,
$\vz_t = \sqrt{\bar\alpha_t}\,\vz_0 + \sqrt{1-\bar\alpha_t}\,\veps$,
$\veps \sim \mathcal{N}(0, \matI)$. The single learned object is a network
$\veps_\theta(\vz_t, t)$ predicting the added noise, equivalently the score
$\vs_\theta(\vz_t, t) = -\veps_\theta(\vz_t, t)/\sqrt{1-\bar\alpha_t} \approx \nabla_{\vz_t}\log p_t(\vz_t)$;
it also gives the clean estimate
$\tilde \vz_0(\vz_t) = (\vz_t + (1-\bar\alpha_t)\,\vs_\theta(\vz_t, t))/\sqrt{\bar\alpha_t}$.
Sampling denoises one step at a time, written $\mathcal{D}_t$, in
either the DDPM ancestral form
\begin{equation}
\mathcal{D}^{\mathrm{DDPM}}_t(\vz_t) \;=\; \tfrac{1}{\sqrt{\alpha_t}}\big(\vz_t + (1-\alpha_t)\,\vs_\theta(\vz_t, t)\big) \;+\; \sigma_t \vxi,
\label{eq:ddpm-step}
\end{equation}
where $\alpha_t = \bar\alpha_t/\bar\alpha_{t-1},\;\; \vxi \sim \mathcal{N}(0, \matI)$ or the DDIM form used at inference \citep{song2020ddim},
\begin{equation}
\mathcal{D}^{\mathrm{DDIM}}_t(\vz_t) \;=\; \sqrt{\bar\alpha_{t-1}}\,\tilde \vz_0(\vz_t)
\;+\; \sqrt{1-\bar\alpha_{t-1}-\sigma_t^2}\,\veps_\theta(\vz_t, t)
\;+\; \sigma_t \vxi,
\label{eq:ddim}
\end{equation}
whose stochasticity $\sigma_t$ is set by a coefficient $\eta_{\ddim} \in [0,1]$
($\eta_{\ddim}{=}0$ deterministic, $\eta_{\ddim}{=}1$ recovering the DDPM
step \eqref{eq:ddpm-step}).

\subsection{ADMM in Scaled Form}
\label{subsec:admm-bg}

ADMM \citep{boyd2011admm} solves the two-block consensus
$\min_{\vx,\vz} F(\vx) + G(\vz)$ s.t.\ $\vx = \vz$ by alternating minimization of
the scaled augmented Lagrangian
$L_\rho(\vx, \vz, \vu) = F(\vx) + G(\vz) + (\rho/2)\norm{\vx - \vz + \vu}^2$ where
$\vu = \vlambda/\rho$ is the scaled dual. Throughout, the
\emph{proximal operator} of a function $h : \R^d \to \R \cup \{+\infty\}$
at penalty $\rho > 0$ is
\begin{equation}
\label{eq:prox-def}
\prox_{h/\rho}(\vv) \;:=\; \argmin_{\vw \in \R^d}\;
  h(\vw) + \tfrac{\rho}{2}\norm{\vw - \vv}^2,
\end{equation}
the point that trades off minimizing $h$ against staying close to the
input $\vv$, with $\rho$ setting the trade-off and larger $\rho$ keeping the
solution nearer $\vv$; when $h$ includes an indicator $\iota_{\mathcal{S}}$
of a set $\mathcal{S}$, the prox enforces $\vw \in \mathcal{S}$ exactly.
The ADMM updates are then
\begin{equation}
\label{eq:admm-updates}
\left\{
\begin{aligned}
\vx^{k+1} &= \prox_{F/\rho}(\vz^k - \vu^k), \\
\vz^{k+1} &= \prox_{G/\rho}(\vx^{k+1} + \vu^k), \\
\vu^{k+1} &= \vu^k + (\vx^{k+1} - \vz^{k+1}).
\end{aligned}
\right.
\end{equation}
The scaled dual $\vu$ accumulates the consensus residual $\vx - \vz$,
adaptively enforcing the constraint.

\section{Method and Theoretical Analysis}
\label{sec:method}

\subsection{Posterior MAP Framing}
\label{subsec:map-framing}
We frame motion planning as a maximum a posteriori (MAP) problem over a
trajectory parameter $\vz \in \R^d$ decoded to a joint-space configuration
path $\vq(\tau; \vz)$, given task context
$y = (\mathcal{O}, \vq^{\mathrm{start}}, \mathcal{Q}^{\mathrm{goal}})$ of
obstacles, start configuration, and a goal that is either a single
configuration or a positive-dimensional $\mathrm{SE}(3)$ end-effector
manifold:
\begin{equation}
\max_{\vz}\; \log p_\phi(\vz \mid \vq^{\mathrm{start}}) \;+\; \log c(\vz, y),
\label{eq:map}
\end{equation}
where $p_\phi$ is the pretrained trajectory diffusion prior (Sec.~\ref{subsec:diffusion})
conditioned only on the start configuration and not on the obstacle field,
and $c(\vz, y) \propto \exp(-J(\vz, y))\,\I[\vz \in \mathcal{C}(y)]$ is the
motion-planning likelihood. Its soft cost
$J(\vz, y) = \sum_i w_i J_i(\vz, y)$ is the standard trajectory-optimization
objective \citep{ratliff2009chomp, schulman2014trajopt, mukadam2018gpmp2,
carvalho2024mpd}; $\mathcal{C}(y)$ is the hard feasibility set induced
by the start, goal, and limit constraints. Direct guidance
\citep{carvalho2024mpd, chung2023dps} attacks \eqref{eq:map} by adding
$\nabla_{\vz} \log c$ to the score $\vs_\theta$ during denoising,
fusing the two responsibilities into one perturbed denoising trajectory.

We instead \emph{decouple} the two log-densities with an auxiliary
variable and solve the resulting consensus program by ADMM
(Sec.~\ref{subsec:admm-bg}): let $\vx$ carry the
diffusion prior and $\vz$ carry the motion-planning cost, tied by a
Euclidean consensus in normalized B-spline coefficient space:
\begin{equation}
\max_{\vx, \vz}\; \log p_\phi(\vx \mid \vq^{\mathrm{start}}) \;+\; \log c(\vz, y)
\qquad \text{s.t.}\;\; \vx = \vz.
\label{eq:topa-problem}
\end{equation}
Setting $F(\vx) := -\log p_\phi(\vx \mid \vq^{\mathrm{start}})$ and
$G(\vz) := -\log c(\vz, y) = J(\vz, y) + \iota_{\mathcal{C}(y)}(\vz)$, this is
the two-block consensus problem of Sec.~\ref{subsec:admm-bg} with scaled
augmented Lagrangian
$L_\rho(\vx, \vz, \vu) = F(\vx) + G(\vz) + (\rho/2)\norm{\vx - \vz + \vu}^2 - (\rho/2)\norm{\vu}^2$,
$\vu = \vlambda/\rho$. Specializing the generic updates \eqref{eq:admm-updates}
to our problem and indexing by denoising timestep $t$ instead of
ADMM iteration count, the per-step \method{} updates are
\begin{equation}
\label{eq:topa-abstract}
\left\{
\begin{aligned}
\vx_{t-1} &= \prox_{F/\rho}\!\big(\vz_t - \vu_t\big)
         = \argmin_{\vx}\; -\log p_\phi(\vx \mid \vq^{\mathrm{start}}) + \tfrac{\rho}{2}\norm{\vx - (\vz_t - \vu_t)}^2, \\
\vz_{t-1} &= \prox_{G/\rho}\!\big(\vx_{t-1} + \vu_t\big)
         = \argmin_{\vz}\; J(\vz, y) + \iota_{\mathcal{C}(y)}(\vz) + \tfrac{\rho}{2}\norm{\vz - (\vx_{t-1} + \vu_t)}^2, \\
\vu_{t-1} &= \vu_t + (\vx_{t-1} - \vz_{t-1}).
\end{aligned}
\right.
\end{equation}
Neither prox in \eqref{eq:topa-abstract} admits a closed form: $F$ is
implicit in the pretrained denoiser $\vs_\theta$, and $G$ is non-convex
on every constraint.
\method{} replaces each prox with a principled inexact update, summarized in
Algorithm~\ref{alg:topa}.

\subsection{\texorpdfstring{The $\vx$-Block}{The x-Block}: Prior Prox as One Denoising Step}
\label{subsec:xblock}
The prior prox $\prox_{F_t/\rho}$ in \eqref{eq:topa-abstract} has no
closed form, since $F_t := -\log p_{\phi,t}(\cdot \mid \vq^{\mathrm{start}})$
is implicit in the denoiser. We approximate it by a single
denoising step $\mathcal{D}_t$ of size $\eta_t$: because the
denoiser's score $\vs_\theta$ estimates the prior score $-\nabla F_t$ up to a
score-estimation error $\ve_t$, such a step is, to first order, a gradient
step on $F_t$ and hence the prior prox, as the following identity makes
precise.

\begin{proposition}[A denoising step is a first-order prior prox]
\label{prop:x-firstorder}
Let $\eta_t := 1-\alpha_t$ and $\rho = 1/\eta_t$. If $F_t$ has locally Lipschitz
gradient near $y$, the prior prox has the first-order expansion
$\prox_{F_t/\rho}(y) = y - \eta_t\nabla F_t(y) + O(\eta_t^2)$ --- a single
gradient step on $F_t$ --- and the DDPM step \eqref{eq:ddpm-step} realizes that
step through its score, so it is a \emph{first-order approximation} of the
prior prox with total error
\begin{equation}
\mathcal{D}^{\mathrm{DDPM}}_t(y) - \prox_{F_t/\rho}(y)
\;=\; \underbrace{\eta_t\,\ve_t(y)}_{\textnormal{(i) score error}}
\;+\; \underbrace{\tfrac12\eta_t\, y}_{\textnormal{(ii) VP rescaling}}
\;+\; \underbrace{O(\eta_t^2)}_{\textnormal{(iii) higher order}}.
\label{eq:x-firstorder}
\end{equation}
\end{proposition}

This identity is what the implemented sampler inherits. At inference we
sample with DDIM \eqref{eq:ddim} --- the same denoising process reparameterized for deterministic or few-step sampling
($\eta_{\ddim}$ interpolating back to DDPM ancestral at
$\eta_{\ddim}{=}1$) --- so the $\vx$-block is one DDIM step from the
dual-shifted noisy state $\hat \vx_t := \vz_t - \vu_t$,
\begin{equation}
\vx_{t-1} \;\approx\; \mathcal{D}^{\mathrm{DDIM}}_t\!\big(\vz_t - \vu_t\big).
\label{eq:x-block}
\end{equation}
Identity~\eqref{eq:x-firstorder} holds exactly only for the idealized step;
the implemented DDIM update discretizes coarsely and may inject
stochasticity. We therefore invoke the weaker \emph{objective-value}
statement of Proposition~\ref{prop:x-block}, which holds for \emph{any}
denoising update and folds the first-order residual, discretization, and
stochasticity into a single quantity.

\begin{proposition}[The denoising sampler is an inexact prior prox]
\label{prop:x-block}
Fix timestep $t$, let $\va_t := \vz_t - \vu_t$, and let
$P_t^x(\vx; \va_t) := F_t(\vx) + (\rho/2)\norm{\vx - \va_t}^2$, whose minimizer
is the exact prior prox $\vx_t^\star := \prox_{F_t/\rho}(\va_t)$. Assume
$P_t^x(\cdot; \va_t)$ is $\mu_t^x$-strongly convex on a neighbourhood
containing the iterate (a sufficient condition is $F_t$ being
$m_t^F$-weakly convex with $\rho > m_t^F$, giving $\mu_t^x = \rho - m_t^F$).
Let $\vx_{t-1} = \mathcal{D}_t(\va_t)$ be the implemented denoising update with
finite proximal stationarity residual
$\delta_t^x := \mathrm{dist}(0,\, \partial F_t(\vx_{t-1}) + \rho(\vx_{t-1} - \va_t))$.
Then the DDIM step $\mathcal{D}_t(\va_t)$ lies within $\delta_t^x/\mu_t^x$ of the
exact prior prox,
\begin{equation}
\norm{\,\mathcal{D}_t(\va_t) \;-\; \prox_{F_t/\rho}(\va_t)\,}
\;\le\; \frac{\delta_t^x}{\mu_t^x},
\label{eq:prop1-result}
\end{equation}
and is an $\varepsilon_t^x$-inexact proximal point in objective value,
$P_t^x(\vx_{t-1}; \va_t) \le P_t^x(\vx_t^\star; \va_t) + \varepsilon_t^x$ with
$\varepsilon_t^x = (\delta_t^x)^2/(2\mu_t^x)$.
\end{proposition}

The exact prox $\prox_{F_t/\rho}(\va_t)$ is not computable ($F_t$ is implicit in
the denoiser), so the $\vx$-block takes one DDIM step instead;
Proposition~\ref{prop:x-block} certifies this is sound, bounding the gap to the
true prox by the residual $\delta_t^x$.

\subsection{\texorpdfstring{The $\vz$-Block}{The z-Block}: Planning Prox by Gradient Descent}
\label{subsec:zblock}
The cost $J$ is defined on clean trajectories,
so $\prox_{G/\rho}$ at the dual-shifted target $\hat \vz_t := \vx_{t-1} + \vu_t$
cannot be evaluated directly on noisy $\vz$. We bridge using the
clean estimate
$\tilde \vz_0(\vz) = (\vz + (1-\bar\alpha_t)\, \vs_\theta(\vz, t)) / \sqrt{\bar\alpha_t}$,
and take the gradient with respect to noisy $\vz$ via a single
vector-Jacobian product through the denoiser,
\begin{equation}
\vg_{\mathrm{cost}}(\vz) \;:=\; \nabla_{\vz}\, J\!\big(\tilde \vz_0(\vz),\, y\big)
\;=\; \Big(\frac{\partial \tilde \vz_0}{\partial \vz}\Big)^{\!\!\top}\!
       \nabla_{\tilde \vz_0} J\!\big(\tilde \vz_0(\vz),\, y\big).
\label{eq:gcost}
\end{equation}
The gradient of the augmented Lagrangian is then
the sum of an ADMM \emph{coupling force} and the \emph{guidance force},
\begin{equation}
\nabla_{\vz} L_\rho(\vz) \;=\; \vF_{\mathrm{coup}}(\vz) + \gamma\, \vg_{\mathrm{cost}}(\vz),
\qquad
\vF_{\mathrm{coup}}(\vz) \;:=\; \rho\,(\vz - \vx_{t-1} - \vu_t),
\label{eq:zgrad}
\end{equation}
and we approximate $\prox_{G/\rho}$ by $K$ steps of plain gradient descent,
\begin{equation}
\vz^{(k+1)} \;=\; \vz^{(k)} - \eta\,\big[\vF_{\mathrm{coup}}(\vz^{(k)}) + \gamma\, \vg_{\mathrm{cost}}(\vz^{(k)})\big],
\quad k = 0, \dots, K{-}1,
\label{eq:zstep}
\end{equation}
with $\vz_{t-1} := \vz^{(K)}$; the scalar cost weight $\gamma$ (the planning
term is $\gamma J$, per-task tuned in Sec.~\ref{subsec:ablations}) balances
coupling against guidance.
The boundary control points
are pinned to the start/goal anchors and the hard feasibility set
$\iota_{\mathcal{C}(y)}$ is enforced as a projection on those indices;
this and the remaining motion-planning implementation choices are deferred
to Appendix~\ref{app:to-formulation}.

\begin{proposition}[Finite-$K$ planning prox is an inexact prox]
\label{prop:z-block}
Fix timestep $t$, let $\vc_t := \vx_{t-1} + \vu_t$ be the planning center,
and define
$Q_t(\vz; \vc_t) := \gamma J_t(\vz; y) + (\rho/2)\norm{\vz - \vc_t}^2$
with $J_t(\vz; y) := J(\tilde \vz_0(\vz), y)$ the cost composed with the
clean-estimate map. Under standard regularity (clean cost $J$ is $L_J$-smooth
and $m_J$-weakly convex on the relevant compact set, clean-estimate map satisfies
$\|D\tilde \vz_0(\vz)\| \le M_t$ and $N_t$-Lipschitz Jacobian), the composite
$J_t$ is $L_t$-smooth and $m_t$-weakly convex with
$L_t \le M_t^2 L_J + B_J N_t$ and $m_t \le M_t^2 m_J + B_J N_t$.
If $\rho > \gamma m_t$, then $Q_t(\cdot; \vc_t)$ is
$\mu_t^z := (\rho - \gamma m_t)$-strongly convex and
$L_t^Q := (\rho + \gamma L_t)$-smooth, so the $K$ gradient-descent steps
\eqref{eq:zstep} with $0 < \eta \le 1/L_t^Q$ converge \emph{geometrically} to
the exact planning prox. Writing
$\Delta_t^{(0)} := Q_t(\vz^{(0)}; \vc_t) - \min_{\vz} Q_t(\vz; \vc_t)$ for the
initial objective gap and $r_t := 1 - \eta\mu_t^z \in [0, 1)$ for the per-step
contraction factor, the output $\vz_{t-1} = \vz^{(K)}$ satisfies
\begin{equation}
\begin{gathered}
\gamma J_t(\vz_{t-1}; y) + \tfrac{\rho}{2}\norm{\vz_{t-1} - \vc_t}^2
\;\le\; \min_{\vz} Q_t(\vz; \vc_t) + \varepsilon_t^z, \\
\varepsilon_t^z \;=\; r_t^{\,K}\,\Delta_t^{(0)} \;=\; (1 - \eta\mu_t^z)^K\,\Delta_t^{(0)}.
\end{gathered}
\label{eq:prop2-result}
\end{equation}
That is, the prox inexactness $\varepsilon_t^z$ decays \emph{geometrically} in
the inner-step count $K$ at rate $r_t = 1 - \eta\mu_t^z < 1$, so a small $K$
already yields a tight inexact prox.
\end{proposition}

Proof and the clean-estimate-Jacobian derivation are in
Appendix~\ref{app:prop-z-block}; a companion bound on how far one
$\vz$-block can move from the prox centre, used by the implementation,
is given in Appendix~\ref{app:to-formulation}.

\begin{algorithm}[t]
\caption{\method{}: ADMM-decoupled denoising.}
\label{alg:topa}
\begin{algorithmic}[1]
\State \textbf{Input:} score $\vs_\theta$; query $y = (\mathcal{O}, \vq^{\mathrm{start}}, \mathcal{Q}^{\mathrm{goal}})$;
   cost $J$; schedule $\{\bar\alpha_t\}_{t=1}^{T}$; penalty $\rho$; inner steps $K$, step size $\eta$, cost scale $\gamma$
\State $\vx_T, \vz_T \sim \mathcal{N}(0, \matI)$; \;\; $\vu_T \leftarrow 0$ \Comment{scaled dual $\vu = \vlambda/\rho$}
\For{$t = T, T{-}1, \dots, 1$}
   \State $\vx_{t-1} \leftarrow \mathcal{D}^{\mathrm{DDIM}}_t(\vz_t - \vu_t)$ \Comment{$\vx$-block: one frozen prior step, Eq.~\eqref{eq:x-block}}
   \State $\vz \leftarrow \vx_{t-1} + \vu_t$ \Comment{$\vz$-block: initialize at the prox centre $\vc_t = \vx_{t-1} + \vu_t$}
   \For{$k = 1, \dots, K$}
     \State $\vF_{\mathrm{coup}} \leftarrow \rho\,(\vz - \vx_{t-1} - \vu_t)$ \Comment{ADMM coupling force, Eq.~\eqref{eq:zgrad}}
     \State $\tilde \vz_0 \leftarrow \big(\vz + (1-\bar\alpha_t)\,\vs_\theta(\vz, t)\big) / \sqrt{\bar\alpha_t}$ \Comment{Tweedie clean estimate}
     \State $\vg_{\mathrm{cost}} \leftarrow (\partial \tilde \vz_0/\partial \vz)^{\!\top}\, \nabla_{\tilde \vz_0} J(\tilde \vz_0, y)$ \Comment{guidance force, Eq.~\eqref{eq:gcost}}
     \State $\vz \leftarrow \vz - \eta\,(\vF_{\mathrm{coup}} + \gamma\, \vg_{\mathrm{cost}})$ \Comment{inner gradient step, Eq.~\eqref{eq:zstep}}
   \EndFor
   \State $\vz_{t-1} \leftarrow \vz$
   \State $\vu_{t-1} \leftarrow \vu_t + (\vx_{t-1} - \vz_{t-1})$ \Comment{scaled dual update, Eq.~\eqref{eq:topa-abstract}}
\EndFor
\State \Return $\vz_0$ (decoded into a B-spline trajectory)
\end{algorithmic}
\end{algorithm}

\subsection{Finite-Horizon Convergence Guarantee}
\label{subsec:convergence}
Algorithm~\ref{alg:topa} summarizes the resulting per-step iteration: an
$\vx$-block denoising step \eqref{eq:x-block}, a $K$-step $\vz$-block descent
\eqref{eq:zstep}, and the scaled dual update of \eqref{eq:topa-abstract}.
For this algorithm, we state a
finite-horizon conditional residual bound that absorbs all
inexactness into measurable quantities. Define the primal residual
$\vr_t := \vx_t - \vz_t$, the $\vz$-movement $\vd_t := \vz_t - \vz_{t+1}$, and the
\emph{nonstationary drift}
\begin{equation}
\begin{aligned}
\omega_t \;:=\; \sup_{\vz \in \mathcal{Z}}\!\big[\,
&|F_{t-1}(\vz) - F_t(\vz)| + \gamma |J_{t-1}(\vz; y) - J_t(\vz; y)| \\
&\quad + \tfrac{1}{\rho}\norm{\gamma\nabla J_{t-1}(\vz; y) - \gamma\nabla J_t(\vz; y)}^2\,\big],
\end{aligned}
\label{eq:drift}
\end{equation}
which measures schedule-induced changes between adjacent denoising steps.

\begin{theorem}[Conditional finite-horizon residual bound]
\label{thm:finite-horizon}
Suppose along the generated trajectory the iterates remain in a compact
set $\mathcal{Z}$ on which the Lagrangian is bounded below, the per-step
prior-prox and planning-prox errors $\varepsilon_t^x, \varepsilon_t^z$ from
Propositions~\ref{prop:x-block} and~\ref{prop:z-block} are finite, the
planning prox is uniformly strongly convex with
$\mu_z := \inf_t (\rho - \gamma m_t) > 0$, and the standard nonconvex-ADMM
dual-residual domination condition (made precise in
Appendix~\ref{app:thm-finite-horizon}) holds. Then there exists a constant
$C > 0$ independent of $T$ such that
\begin{equation}
\min_{0 \le t < T}\big(\norm{\vr_t}^2 + \norm{\vd_t}^2\big)
\;\le\; \frac{C}{T}\!\left[
\mathcal{L}_{\rho,T}(\vx_T, \vz_T, \vu_T) - \underline{\mathcal{L}}
+ \sum_{t=1}^T \big(\varepsilon_t^x + \varepsilon_t^z + \omega_t\big)
\right].
\label{eq:thm3-result}
\end{equation}
\end{theorem}

\eqref{eq:thm3-result} is conditional: it states that under controlled
per-step inexactness ($\varepsilon_t^x, \varepsilon_t^z$) and bounded
schedule drift ($\omega_t$), the minimum primal-and-movement residual
over the denoising schedule decays at rate $C/T$. The bound is not a global
optimality claim and not an asymptotic convergence result — it is the
right level of guarantee for a finite-horizon sampler with a neural
denoising step, a fixed-$K$ planning inner loop, and a changing diffusion
schedule. A sufficient condition for the dual-residual domination via the
planning-gradient Lipschitz constant, and the telescoping proof, are in
Appendix~\ref{app:thm-finite-horizon}.

\section{Experiments}
\label{sec:expts}

\label{subsec:setup}
We evaluate on six benchmark tasks \citep{carvalho2024mpd}: point-mass \textbf{simple2d} and
\textbf{narrowpass} ($\R^2$, joint goal), \textbf{planar2link} (2-DoF, joint
goal), \textbf{planar4link} (4-DoF, $\mathrm{SE}(2)$ goal, 1-D redundancy), and
7-DoF Franka Panda \textbf{spheres3d} and \textbf{warehouse} ($\mathrm{SE}(3)$
goal, 1-D redundancy). We compare three samplers under identical seeds and
DDIM schedule --- \textit{vanilla} (DDIM, no guidance), \textit{MPD}
\citep{carvalho2024mpd}, and \method{}. The comparison is \emph{fair}: all
three share the same pretrained prior, cost, DDIM schedule, and seeds, and MPD
and \method{} further share the same motion-planning configuration --- identical
start/goal anchoring and, on the redundant-arm tasks, the same prior-anchored
goal-manifold post-process (App.~\ref{app:to-formulation}). MPD is run in its
standard \emph{clean-state} form, so the sole methodological difference is the sampler --- MPD's direct guidance 
vs.\ \method{}'s ADMM decoupling mechanism. Metrics:
fraction-valid \textbf{FV}, collision intensity \textbf{CI},
end-effector position / orientation error
$\mathbf{ee}_{\mathrm{pos}}$~(m)\,/\,$\mathbf{ee}_{\mathrm{ori}}$~(deg) on
$\mathrm{SE}$-goal tasks, and prior preservation
$D_{\mathrm{prior}} := \E_b\!\norm{\tau_b - \tau_b^{\mathrm{prior}}}_2$. Each
table cell aggregates 5 seeds $\times$ 10 instances $\times$ 100 samples
$=5{,}000$ trajectories; we use $T_{\ddim}=15$ at $\eta_{\ddim}=0.5$ on the four narrower-corridor
tasks and $\eta_{\ddim}=1.0$ (full DDPM) on the two Panda tasks, identical for
all algorithms.

\subsection{Cross-Task Benchmark}
\label{subsec:main-results}

Table~\ref{tab:main} reports feasibility, prior preservation, and end-effector
accuracy across all six tasks; every $\star$ marks a paired 2$\sigma$
significance against MPD. \emph{\method{} strictly Pareto-dominates MPD on FV,
CI, $\mathrm{ee}_{\mathrm{pos}}$, $\mathrm{ee}_{\mathrm{ori}}$, and
$D_{\mathrm{prior}}$ on 4/6 tasks---narrowpass, planar2link, planar4link,
spheres3d---with significance on both FV and $D_{\mathrm{prior}}$; on the
remaining two it does not dominate but shifts the feasibility--fidelity
trade-off}: on warehouse it raises FV and lowers CI at $D_{\mathrm{prior}}$
within seed noise, whereas on simple2d it gives up FV (0.96 vs.\ 0.99) and CI
(4.0 vs.\ 0.1) for a smaller $D_{\mathrm{prior}}$, on a task where MPD is
already saturated (FV 0.99). The dominating
axis on redundant arms is orientation---across planar4link, spheres3d, and
warehouse, $\mathrm{ee}_{\mathrm{ori}}$ collapses from MPD's
$1.39^\circ$--$2.14^\circ$ to \method{}'s $0.08^\circ$--$0.37^\circ$, a
\textbf{6--17$\times$ reduction}. Both methods run the \emph{same}
prior-anchored goal-manifold post-process (Eq.~\eqref{eq:manifold-prox}), which
slides the terminal configuration along the EE-pose null space to the
closest-to-prior pose meeting the $\mathrm{SE}(3)$ goal; the gap therefore
reflects the terminal configurations \method{}'s decoupled sampler feeds that
shared prox, which it refines to near-zero orientation error. The
same ranking holds on the (FV, $D_{\mathrm{prior}}$) Pareto frontier
(Appendix Fig.~\ref{fig:pareto}): \method{} occupies the Pareto-favorable upper-right
quadrant on the four dominated tasks, trades 3\,pp FV for a $D_{\mathrm{prior}}$
win on simple2d, and matches MPD's $D_{\mathrm{prior}}$ while improving FV on
warehouse. On most tasks these gains arrive \emph{together}: \method{} improves
feasibility (FV\,$\uparrow$, CI\,$\downarrow$) \emph{while} reducing the prior
deviation $D_{\mathrm{prior}}$, rather than bartering one against the other as a
single scalar guidance weight must---the behaviour the ADMM decoupling is built
for, with the $\vx$-block holding samples on the learned diffusion manifold and
the $\vz$-block repairing constraints. simple2d is the exception: there MPD is
already near-saturated (FV 0.99, CI 0.1) and \method{} instead trades a little
feasibility for a closer-to-prior solution. Appendix
Fig.~\ref{fig:traj-overlay} makes this concrete per sample on the harder tasks:
\method{}'s trajectories form a broad, prior-like bundle that keeps the prior's
coverage yet is almost entirely valid (collision-free, blue), whereas the
unguided prior is equally broad but riddled with collisions (red) and direct
guidance clears fewer of them.

\begin{table}[t]
\centering
\caption{Main results on the six tasks. \textbf{FV} (fraction valid,
$\uparrow$): fraction of trajectories that are collision-free,
dynamic-limit-feasible, and goal-reaching; \textbf{CI} (collision intensity,
$\downarrow$): mean signed-distance penetration; $D_{\mathrm{prior}}$
($\downarrow$): distance to the unguided prior sample, so smaller means closer
to the learned prior; $\mathrm{ee}_{\mathrm{pos}}$\,/\,$\mathrm{ee}_{\mathrm{ori}}$
($\downarrow$): terminal end-effector position\,/\,orientation error, defined
only on the three redundant-arm $\mathrm{SE}$-goal tasks (``--'' elsewhere).
Arrows mark the favourable direction; \textbf{bold} = best per task and metric,
$\star$ = a significant improvement over MPD, and vanilla is the
$D_{\mathrm{prior}}$ reference. \method{} improves the feasibility--fidelity
trade-off on most tasks; the exception is simple2d, where it gives up FV and CI
for a smaller $D_{\mathrm{prior}}$.}
\label{tab:main}
\small
\setlength{\tabcolsep}{3.5pt}
\resizebox{\linewidth}{!}{%
\begin{tabular}{l ccc ccc cc cc cc}
\toprule
 & \multicolumn{3}{c}{FV $\uparrow$} & \multicolumn{3}{c}{CI $\downarrow$ ($\times 10^{-3}$)} & \multicolumn{2}{c}{$D_{\mathrm{prior}}$ $\downarrow$} & \multicolumn{2}{c}{$\mathrm{ee}_{\mathrm{pos}}$ $\downarrow$ (m)} & \multicolumn{2}{c}{$\mathrm{ee}_{\mathrm{ori}}$ $\downarrow$ ($^\circ$)} \\
\cmidrule(lr){2-4} \cmidrule(lr){5-7} \cmidrule(lr){8-9} \cmidrule(lr){10-11} \cmidrule(lr){12-13}
task & van. & MPD & \textbf{TOPA} & van. & MPD & \textbf{TOPA} & MPD & \textbf{TOPA} & MPD & \textbf{TOPA} & MPD & \textbf{TOPA} \\
\midrule
simple2d     & 0.78 & \textbf{0.99} & 0.96                 & 24 & \textbf{0.1} & 4.0            & .044 & \textbf{.029}$^\star$ & -- & -- & -- & -- \\
narrowpass   & 0.65 & 0.87          & \textbf{0.91}$^\star$ & 22 & 3.1 & \textbf{0.5}$^\star$  & .050 & \textbf{.044}$^\star$ & -- & -- & -- & -- \\
planar2link  & 0.63 & 0.74          & \textbf{0.96}$^\star$ & 40 & 20 & \textbf{1.8}$^\star$   & .163 & \textbf{.107}$^\star$ & -- & -- & -- & -- \\
planar4link  & 0.18 & 0.61          & \textbf{0.79}$^\star$ & 139 & 23 & \textbf{11}$^\star$   & .360 & \textbf{.222}$^\star$ & .027 & \textbf{.012}$^\star$ & 1.39 & \textbf{0.08}$^\star$ \\
spheres3d    & 0.41 & 0.72          & \textbf{0.82}$^\star$ & 76 & 10 & \textbf{8.6}            & .622 & \textbf{.600}$^\star$ & .026 & \textbf{.011}$^\star$ & 2.14 & \textbf{0.37}$^\star$ \\
warehouse    & 0.64 & 0.84          & \textbf{0.88}$^\star$ & 35 & 3.7 & \textbf{1.9}$^\star$  & \textbf{.481} & .498 & .023 & \textbf{.010}$^\star$ & 1.78 & \textbf{0.27}$^\star$ \\
\bottomrule
\end{tabular}%
}
\end{table}

\subsection{Ablations and sensitivity}
\label{subsec:ablations}

We ablate two core components on planar4link (a
representative redundant-arm task and the largest \method{}-over-MPD gap); both
are 5-seed verified at $2\sigma$. \emph{Goal-manifold post-process:} removing it
reverts the terminal orientation error $\mathrm{ee}_{\mathrm{ori}}$ from
$0.08^\circ$ to $4.67^\circ$ (a factor of $57\times$), with FV
($0.80\!\to\!0.74$) and $D_{\mathrm{prior}}$ ($0.22\!\to\!0.21$) little
changed---a cheap terminal refinement ($30\%$ wall-time) that buys the
$\mathrm{SE}$-goal accuracy. \emph{Full-gradient VJP:} replacing the VJP through
the denoiser with the stop-grad approximation
$\partial s_\theta/\partial z \approx 0$ inflates the noise-space guidance by
$1/\sqrt{\bar\alpha_t}$ (up to $60\times$ on a 1000-step schedule), dropping FV
by 13\,pp ($0.80\!\to\!0.67$) and inflating $D_{\mathrm{prior}}$ from $0.22$ to
$0.34$ without per-task $\gamma$ re-tuning---the manifold-projected
cost gradient, not a raw one, is what keeps corrections on the learned manifold.
\emph{Tweedie vs.\ direct gradient.} A param-matched comparison (seed~2)
replaces the noise-state Tweedie/VJP gradient with a \emph{direct} cost gradient
on the control points, holding all ADMM/cost/schedule knobs fixed. This inflates $D_{\mathrm{prior}}$ by
$2$--$2.5\times$ ($0.21\!\to\!0.52$ on planar4link, $0.047\!\to\!0.089$ on
narrowpass) and lowers FV by ${\sim}10$\,pp, \emph{despite} a perturbation cap
that actively limits drift. The denoiser-projected gradient is
thus what holds samples near the prior, consistent with \method{}'s smaller
$D_{\mathrm{prior}}$ than the clean-state MPD baseline.

Appendix Fig.~\ref{fig:heatmap} reports the per-task FV heatmap over
$\rho \in \{6, 9, 11, 13, 15, 18\}$ and $\gamma \in \{0.5, 0.7,
0.85, 1.0\}$ at $K = 4$ (3 seeds per cell, $24 \times 6 = 144$ runs), and three
regularities emerge. \textbf{Two basins:} most tasks prefer $\rho \in [6, 9]$
at $\gamma \in [0.7, 1.0]$ (simple2d, narrowpass, planar2link,
spheres3d, warehouse), while planar4link prefers the stiffer $\rho = 11$ at
$\gamma = 0.85$, where the manifold prox absorbs the terminal
slack; joint-space and planar tasks favour weaker coupling because the cost
gradient has more voice at smaller $\rho$. \textbf{spheres3d is
$\gamma$-sensitive:} its random-sphere clutter concentrates the
collision gradient, so full-strength $\gamma = 1.0$ over-corrects
and inflates $D_{\mathrm{prior}}$ by 13\,\%, whereas halving $\gamma$
both raises FV and lowers $D_{\mathrm{prior}}$; the other five tasks are robust
across $\gamma \in [0.7, 1.0]$. \textbf{Wide basin:} for all six
tasks the optimum is not knife-edge---a $2 \times 2$ neighborhood stays within
2\,pp of the best FV and 5\,\% of the best $D_{\mathrm{prior}}$. \method{} is
therefore \emph{parameter-robust}: the per-task knobs of
Sec.~\ref{subsec:setup} are interpretable choices inside a wide flat basin, not
fragile optima.

\section{Limitations}
\label{sec:limitations}

First, TOPA does not provide a per-sample hard feasibility guarantee. 
Its ADMM split is designed to decouple prior preservation from constraint satisfaction,
but the planning branch remains an inexact finite-step. 
Therefore, for safety- or deployment-critical settings, TOPA should be treated as a proposal generator 
and used with multiple trajectory samples followed by feasibility checking.
Second, TOPA introduces additional inference-time computation.
At each active denoising step, it estimates a clean trajectory and backpropagates 
the planning cost before performing the inner planning updates, 
which is more expensive than vanilla diffusion or direct clean-state guidance.
This overhead is the price of applying geometry- and dynamics-aware corrections in a space 
that better preserves the learned trajectory prior.

\section{Conclusion}
\label{sec:conclusion}

We presented TOPA, a training-free motion-planning diffusion framework that decouples prior preservation 
and constraint satisfaction through an ADMM-style split between a diffusion-prior trajectory and 
a trajectory-optimization branch. By performing structured clean-state planning updates while maintaining 
consensus with the pretrained diffusion prior, TOPA provides a more effective feasibility--fidelity trade-off
 than direct scalar-weighted guidance. Across navigation, planar manipulation, and high-DoF robot-arm planning tasks, 
 TOPA improves feasibility, collision behavior, end-effector accuracy, and prior preservation on most benchmarks, 
 with particularly strong gains on redundant-arm problems. These results suggest that coupling generative trajectory priors 
 with optimization-structured inference is a promising direction for constraint-aware robot motion planning.

\bibliography{references}

\newpage
\appendix

\section{Trajectory-optimization formulation and implementation}
\label{app:to-formulation}

This appendix expands the trajectory-optimization setup summarized in
Sec.~\ref{sec:method} and collects the motion-planning--specific
realizations of the ADMM blocks deferred from the main text. Trajectories are
joint-space B-splines; the optimization (and diffusion) variable is the vector
of \emph{free} control points $\vz \in \R^d$, while the boundary control points
are pinned to the start (and, on joint-goal tasks, the goal) configuration
through a fixed offset $\vb$ (start/goal anchoring below). We write
$\vq(\tau; \vz)$ for the decoded configuration at time $\tau \in [0,
T_{\mathrm{tot}}]$; it --- and hence velocity, acceleration, and every cost
term --- is linear in $\vz$. Below we give the constrained problem and its soft-penalty
relaxation, then start/goal anchoring, block initialization, and the
goal-manifold post-process.

\paragraph{Constrained problem.}
Given task context
$y = (\mathcal{O}, \vq^{\mathrm{start}}, \mathcal{Q}^{\mathrm{goal}})$ the
motion-planning problem is
\begin{equation}
\begin{aligned}
\min_{\vz \in \R^d}\; & J_{\mathrm{smooth}}(\vz)
\quad \text{s.t.}\;
\vq(\tau; \vz) \in \mathcal{Q}_{\mathrm{free}}(\mathcal{O}),\;
\vq^{\min} \le \vq(\tau; \vz) \le \vq^{\max},\\
& |\dot \vq(\tau; \vz)| \le \dot \vq^{\max},\;
|\ddot \vq(\tau; \vz)| \le \ddot \vq^{\max},
\;\;\forall \tau \in [0, T_{\mathrm{tot}}],\\
& \vq(0; \vz) = \vq^{\mathrm{start}},\quad
\vq(T_{\mathrm{tot}}; \vz) \in \mathcal{Q}^{\mathrm{goal}},
\end{aligned}
\label{eq:traj-opt}
\end{equation}
where the goal set $\mathcal{Q}^{\mathrm{goal}}$ is either a singleton (joint
goal, absorbed into $\vb$) or, for an $\mathrm{SE}(3)$ end-effector goal, the
$(d_q{-}6)$-dimensional submanifold
$\{\vq : T_{ee}(\vq) = T_{ee}^{\mathrm{goal}}\}$ (1-D on the 7-DoF Panda, 2-D
on the 4-link planar arm).

\paragraph{Soft-penalty objective as implemented.}
Every hard constraint is relaxed into a single weighted-sum objective,
evaluated by trapezoidal quadrature over the normalized phase
$s = \tau/T_{\mathrm{tot}} \in [0,1]$,
\begin{equation}
J(\vz, y) \;=\; \sum_{k} w_k \int_0^1 C_k\big(\vq(s; \vz)\big)\, ds,
\label{eq:soft-objective}
\end{equation}
with weights $(w_{\mathrm{coll}}, w_{\mathrm{lim}}, w_{\mathrm{vel}},
w_{\mathrm{acc}})$ inherited from MPD (e.g.\ $(10,\, 1,\, 10^{-2},\, 10^{-3})$);
the velocity and acceleration terms form $J_{\mathrm{smooth}}$. Each $C_k$ is
evaluated on the decoded configuration $\vq(s; \vz)$ and its derivatives.
For the robot's collision spheres $\ell$ of radius $r_\ell$ with world
positions $\vx_\ell(\vq)$ obtained by forward kinematics, and the
signed-distance field $\operatorname{SDF}_{\mathcal{O}}$ of the obstacle set
$\mathcal{O}$ ($\operatorname{SDF}_{\mathcal{O}} > 0$ outside obstacles), the
collision term is the total clamped penetration
\begin{equation}
C_{\mathrm{coll}}(\vq) \;=\; \sum_{\ell} \max\!\big(0,\; (r_\ell + \kappa) - \operatorname{SDF}_{\mathcal{O}}(\vx_\ell(\vq))\big),
\label{eq:cost-coll}
\end{equation}
with a small cutoff margin $\kappa$; it is zero iff every sphere clears every
obstacle. The joint- and dynamic-limit term is a one-sided hinge with safety
margin $\epsilon = 0.03$, active only near a violated bound,
\begin{equation}
C_{\mathrm{lim}}(\vq) \;=\; \tfrac12\big\|\operatorname{ReLU}(\vq^{\min}{+}\epsilon - \vq)\big\|_2
\;+\; \tfrac12\big\|\operatorname{ReLU}(\vq - (\vq^{\max}{-}\epsilon))\big\|_2,
\label{eq:cost-lim}
\end{equation}
with analogous hinges on $\dot\vq$ and $\ddot\vq$ against their limits. The
smoothness regularizers are the velocity and acceleration norms
$C_{\mathrm{vel}}(\vq) = \tfrac12\|\dot\vq\|_2$ and
$C_{\mathrm{acc}}(\vq) = \tfrac12\|\ddot\vq\|_2$, which the diffusion prior
already partly supplies from smooth demonstrations. The analytic gradient
$\vg(\vz, y) = \nabla_{\vz} J(\vz, y)$ backpropagates through the
forward-kinematics Jacobian $\partial \vx/\partial \vq$ and the B-spline map
$\partial \vq/\partial \vz$; the hard feasibility set $\mathcal{C}(y)$ used in
the split of Sec.~\ref{sec:method} is the $w_k \to \infty$ limit of these
penalties, each replaced by the indicator of its constraint.

\paragraph{Start/goal anchoring and hard conditioning.}
The start configuration $\vq^{\mathrm{start}}$ and, on joint-goal tasks, the
goal $\vq^{\mathrm{goal}}_{\mathrm{joint}}$ are pinned exactly by construction:
their B-spline control points live in the fixed offset $\vb$ and are excluded
from the free coordinates $\vz$.
For a $C^2$ clamped B-spline with zero boundary velocity and acceleration this
pins the first and last few control points; write $H$ for this set of
hard-conditioned indices and $\mathrm{apply}_H$ for the projection that resets
them to their anchor values after each block update. Consequently the
consensus constraint $\vx = \vz$ holds exactly on $H$, and the hard
feasibility indicator $\iota_{\mathcal{C}(y)}$ of Sec.~\ref{sec:method} acts as
this projection rather than as a soft penalty. Because $\mathrm{apply}_H$
already enforces $\vx = \vz$ on $H$, the scaled dual carries no residual there
and is masked, $\vu_{t-1}[j] \leftarrow 0$ for $j \in H$, after the dual update
of Algorithm~\ref{alg:topa}.

\paragraph{Block initialization and warm-up gate.}
The two trajectories share a single Gaussian draw, $\vz_T \leftarrow \vx_T$
with $\vx_T = \mathrm{apply}_H(\vxi)$, $\vxi \sim \mathcal{N}(0, \matI)$, which
removes a transient consensus residual at high noise. Because the clean
estimate $\tilde \vz_0$ is too crude to give a useful cost gradient early in
sampling, a warm-up gate runs plain DDIM with $\vu \leftarrow 0$ while
$t / T > f_{\mathrm{start}}$ and activates the $\vz$-block only once
$t / T \le f_{\mathrm{start}}$. Within the active window each $\vz$-prox is
initialized at the current dual-shifted prox centre
$\vz^{(0)} = \mathrm{apply}_H(\vx_{t-1} + \vu_t)$ rather than carried over from
$\vz_t$: on the short schedules we use ($T_{\ddim} = 15$), $\vx_{t-1}$ has just
moved by a full DDIM step, so the carried-over iterate is stale for the
current centre, and starting at the centre devotes all $K$ inner steps to cost
descent. The deviation this introduces is bounded next.

\paragraph{Initialization trust-region bound.}
With the inner loop initialized at the prox centre
$\vz^{(0)} = \vc_t := \vx_{t-1} + \vu_t$, write
$Q_t(\vz; \vc_t) := \gamma J_t(\vz; y) + (\rho/2)\norm{\vz - \vc_t}^2$ for the
planning-prox objective ($J_t(\vz; y) := J(\tilde \vz_0(\vz), y)$, with cost
weight $\gamma$). If $J_t \ge 0$, the $K$ monotone descent steps
\eqref{eq:zstep} satisfy
\begin{equation}
\norm{\vz_{t-1} - \vc_t}^2 \;\le\; \frac{2\gamma}{\rho}\, J_t(\vc_t; y),
\label{eq:init-trustregion}
\end{equation}
a trust-region bound on how far one $\vz$-block can move from the ADMM target.
\emph{Proof.} Gradient descent monotonically decreases $Q_t$, so
$Q_t(\vz^{(K)}; \vc_t) \le Q_t(\vc_t; \vc_t) = \gamma J_t(\vc_t; y)$; combining
with
$Q_t(\vz^{(K)}; \vc_t) = \gamma J_t(\vz^{(K)}; y) + (\rho/2)\norm{\vz^{(K)} - \vc_t}^2$
and $J_t \ge 0$ bounds the quadratic term by $\gamma J_t(\vc_t; y)$.
\eqref{eq:init-trustregion} is small when either the prox centre is already
near-feasible ($J_t(\vc_t; y)$ small) or the coupling is tight ($\rho$ large).

\paragraph{Goal-manifold post-process for redundant arms.}
When $\mathcal{Q}^{\mathrm{goal}}$ in \eqref{eq:traj-opt} is a
positive-dimensional $\mathrm{SE}(3)$ submanifold --- as on the planar 4-link
and 7-DoF Panda --- hard-pinning the terminal control point to a single
witness joint configuration destroys useful null-space behavior the prior has
learned. After the main denoising loop, the guided planners (MPD and
\method{}) run a short prior-anchored prox on the trailing
$n_{\mathrm{term}} = 3$ control points,
\begin{equation}
\vz_0^{\mathrm{trail}} \;\leftarrow\; \argmin_{\vz^{\mathrm{trail}}}\;
   w_g\, J(\vz; y) \;+\; \tfrac{1}{2}\norm{\vz^{\mathrm{trail}} - \vz_0^{\mathrm{trail, prior}}}^2,
\label{eq:manifold-prox}
\end{equation}
solved by $n_M = 20$ steps of clipped (clip $c_M = 0.3$) gradient descent at
step size $\eta_M = 0.03$ with the same $\mathrm{apply}_H$ mask at every step.
The anchor $\vz_0^{\mathrm{trail, prior}}$ is the unguided prior's trailing
control points, so \eqref{eq:manifold-prox} biases the in-manifold choice
toward the prior while the cost gradient drives task-space pose satisfaction.
Applied identically to MPD and \method{}, it sharply lowers orientation error
for both; the 6--17$\times$ \method{}-over-MPD gap in Sec.~\ref{sec:expts} comes
from the better terminal configurations the decoupled sampler feeds this shared
prox, not from the post-process itself.

\section{Proofs for the \texorpdfstring{$\vx$}{x}-block: first-order identity and inexact prox}
\label{app:prop-x-block}

This appendix proves the two $\vx$-block statements of
Sec.~\ref{sec:method} in the order they appear there:
Proposition~\ref{prop:x-firstorder} (a denoising step is a
first-order prior prox) and Proposition~\ref{prop:x-block} (the
implemented denoising step is an inexact prox in objective value).

\subsection*{Proof of Proposition~\ref{prop:x-firstorder}}

\begin{proof}  %
Fix the denoising step $t$ and an input $y$; the only regularity used is that
$F_t := -\log p_{\phi,t}(\cdot \mid \vq^{\mathrm{start}})$ has locally Lipschitz
gradient near $y$. We prove \eqref{eq:x-firstorder} by expanding both sides and
subtracting.

\emph{Prox side.} Let $\vx^+ = \prox_{F_t/\rho}(y) = \argmin_{\vx} F_t(\vx) + \tfrac{\rho}{2}\norm{\vx-y}^2$
with $\rho = 1/\eta_t$; its optimality condition is
$\nabla F_t(\vx^+) + \rho(\vx^+ - y) = 0$. Using the Lipschitz-gradient remainder
$\nabla F_t(\vx^+) = \nabla F_t(y) + O(\norm{\vx^+-y})$ and
$\vx^+ - y = O(\rho^{-1})$,
\begin{align*}
\vx^+
&= y - \rho^{-1}\nabla F_t(\vx^+) \\
&= y - \rho^{-1}\nabla F_t(y) + O(\rho^{-2}) \\
&= y - \eta_t\nabla F_t(y) + O(\eta_t^2),
\end{align*}
the last line using $\rho^{-1} = \eta_t$ --- a single gradient step on $F_t$.

\emph{Denoising side.} Write $\eta_t := 1-\alpha_t$,
$1/\sqrt{\alpha_t} = 1 + \tfrac12\eta_t + O(\eta_t^2)$, and \emph{define} the
score-estimation error $\ve_t(y) := \vs_\theta(y, t) + \nabla F_t(y)$ (so
$\vs_\theta(y, t) = -\nabla F_t(y) + \ve_t(y)$ identically). The deterministic
part of the DDPM step \eqref{eq:ddpm-step} then expands as
\begin{align*}
\mathcal{D}^{\mathrm{DDPM}}_t(y)
&= y + \eta_t\, \vs_\theta(y, t) + \tfrac12\eta_t\, y + O(\eta_t^2) \\
&= y - \eta_t\nabla F_t(y) + \eta_t\, \ve_t(y) + \tfrac12\eta_t\, y + O(\eta_t^2).
\end{align*}

\emph{Subtraction.} The shared $y$ and $-\eta_t\nabla F_t(y)$ terms cancel,
leaving exactly \eqref{eq:x-firstorder},
\[
\mathcal{D}^{\mathrm{DDPM}}_t(y) - \prox_{F_t/\rho}(y)
= \underbrace{\eta_t\, \ve_t(y)}_{\textnormal{(i)}}
+ \underbrace{\tfrac12\eta_t\, y}_{\textnormal{(ii)}}
+ \underbrace{O(\eta_t^2)}_{\textnormal{(iii)}},
\]
where (i) is the denoiser's score-estimation error $\ve_t$ scaled by the step
size $\eta_t$ (zero for an exact score), (ii) is the variance-preserving
rescaling --- deterministic, independent of $F_t$, common to every denoising step,
and dropped in the prior-driven comparison --- and (iii) is the Taylor-truncation
remainder, vanishing as $\eta_t = 1-\alpha_t \to 0$. In particular the proximal stationarity
residual of the denoising step is
$\delta_t^x = \norm{\nabla F_t(\mathcal{D}_t(y)) + \rho(\mathcal{D}_t(y) - y)} = O(\norm{\ve_t} + \eta_t)$.
\end{proof}

\subsection*{Proof of Proposition~\ref{prop:x-block}}

\begin{proof}
Write $\vx_{t-1} = \mathcal{D}_t(\va_t)$ for the implemented step and
$\vx_t^\star = \prox_{F_t/\rho}(\va_t)$ for the exact prox, the minimizer of
$P_t^x(\cdot; \va_t)$. We use only what the proposition states: the
$\mu_t^x$-strong convexity of $P_t^x(\cdot; \va_t)$ on a neighbourhood of both
points, and the stationarity residual $\delta_t^x$, which equals
$\mathrm{dist}(0,\, \partial P_t^x(\vx_{t-1}; \va_t))$ since
$\partial P_t^x(\vx_{t-1}; \va_t) = \partial F_t(\vx_{t-1}) + \rho(\vx_{t-1} - \va_t)$.

\emph{Objective value.} By $\mu_t^x$-strong convexity, for every subgradient
$\vg \in \partial P_t^x(\vx_{t-1}; \va_t)$,
\[
P_t^x(\vx_t^\star; \va_t) \;\ge\; P_t^x(\vx_{t-1}; \va_t)
+ \langle \vg,\, \vx_t^\star - \vx_{t-1}\rangle
+ \tfrac{\mu_t^x}{2}\norm{\vx_t^\star - \vx_{t-1}}^2.
\]
Rearranging, maximizing over the displacement, and minimizing over $\vg$ yields
the quadratic-growth bound
\[
P_t^x(\vx_{t-1}; \va_t) - P_t^x(\vx_t^\star; \va_t)
\;\le\; \frac{\mathrm{dist}\!\big(0,\, \partial P_t^x(\vx_{t-1}; \va_t)\big)^2}{2\mu_t^x}
\;=\; \frac{(\delta_t^x)^2}{2\mu_t^x} \;=:\; \varepsilon_t^x,
\]
which is the objective-value statement of Proposition~\ref{prop:x-block}.

\emph{Distance.} Strong convexity at the minimizer $\vx_t^\star$ gives
$\tfrac{\mu_t^x}{2}\norm{\vx_{t-1} - \vx_t^\star}^2 \le P_t^x(\vx_{t-1}; \va_t) - P_t^x(\vx_t^\star; \va_t) \le \varepsilon_t^x$,
so
\[
\norm{\,\mathcal{D}_t(\va_t) - \prox_{F_t/\rho}(\va_t)\,}
\;=\; \norm{\vx_{t-1} - \vx_t^\star}
\;\le\; \sqrt{2\varepsilon_t^x/\mu_t^x}
\;=\; \frac{\delta_t^x}{\mu_t^x},
\]
which is \eqref{eq:prop1-result}.
\end{proof}

\paragraph{Remark.}
$\delta_t^x$ is computable at run time: $\partial F_t(\vx_{t-1}) = -\vs_\theta(\vx_{t-1}, t)$
up to the score-network approximation, so
$\delta_t^x \approx \norm{-\vs_\theta(\vx_{t-1}, t) + \rho(\vx_{t-1} - \va_t)}$
requires one extra forward pass through the denoiser per denoising step.

\section{Proof of Proposition~\ref{prop:z-block}: finite-$K$ planning prox}
\label{app:prop-z-block}

\begin{proof}
By Proposition~\ref{prop:z-block} the planning-prox objective
$Q_t(\vz; \vc_t) = \gamma J_t(\vz; y) + \tfrac{\rho}{2}\norm{\vz - \vc_t}^2$ is
$\mu_t^z$-strongly convex and $L_t^Q$-smooth, and the steps \eqref{eq:zstep}
are gradient descent on $Q_t$; write $\vz_t^\star := \argmin_{\vz} Q_t(\vz; \vc_t)$.
We carry the per-step decrease into the geometric form \eqref{eq:prop2-result}.

\emph{One-step contraction.} The descent lemma for the $L_t^Q$-smooth $Q_t$ at
step size $0 < \eta \le 1/L_t^Q$ gives
$Q_t(\vz^{(k+1)}) \le Q_t(\vz^{(k)}) - \tfrac{\eta}{2}\norm{\nabla Q_t(\vz^{(k)})}^2$,
while $\mu_t^z$-strong convexity gives the Polyak--\L{}ojasiewicz inequality
$\norm{\nabla Q_t(\vz^{(k)})}^2 \ge 2\mu_t^z\big[Q_t(\vz^{(k)}) - Q_t(\vz_t^\star)\big]$.
Subtracting $Q_t(\vz_t^\star)$ and combining the two,
\[
Q_t(\vz^{(k+1)}) - Q_t(\vz_t^\star)
\;\le\; (1 - \eta\mu_t^z)\big[Q_t(\vz^{(k)}) - Q_t(\vz_t^\star)\big].
\]

\emph{Iterating from $\vz^{(0)}$.} With $r_t := 1 - \eta\mu_t^z \in [0,1)$ and
$\Delta_t^{(0)} := Q_t(\vz^{(0)}; \vc_t) - Q_t(\vz_t^\star; \vc_t)$, composing this
contraction $K$ times and using $\vz_{t-1} = \vz^{(K)}$,
\[
Q_t(\vz_{t-1}; \vc_t) - Q_t(\vz_t^\star; \vc_t)
\;\le\; r_t^{\,K}\,\Delta_t^{(0)} \;=:\; \varepsilon_t^z.
\]

\emph{In the stated form.} Substituting
$Q_t(\vz_{t-1}; \vc_t) = \gamma J_t(\vz_{t-1}; y) + \tfrac{\rho}{2}\norm{\vz_{t-1} - \vc_t}^2$
and $Q_t(\vz_t^\star; \vc_t) = \min_{\vz} Q_t(\vz; \vc_t)$ yields
\[
\gamma J_t(\vz_{t-1}; y) + \tfrac{\rho}{2}\norm{\vz_{t-1} - \vc_t}^2
\;\le\; \min_{\vz} Q_t(\vz; \vc_t) + \varepsilon_t^z,
\qquad
\varepsilon_t^z = r_t^{\,K}\,\Delta_t^{(0)} = (1 - \eta\mu_t^z)^K\,\Delta_t^{(0)},
\]
which is exactly \eqref{eq:prop2-result}.
\end{proof}

\noindent The companion initialization trust-region bound
\eqref{eq:init-trustregion}, used by the implementation, is stated and proved
in Appendix~\ref{app:to-formulation}.

\section{Proof of Theorem~\ref{thm:finite-horizon}: conditional finite-horizon residual bound}
\label{app:thm-finite-horizon}

\begin{proof}
We establish a one-step Lyapunov decrease on $\mathcal{L}_{\rho,t}$ and
telescope it over the finite denoising schedule; throughout, $\mu_z > 0$ is
the uniform planning-prox strong-convexity modulus of
Theorem~\ref{thm:finite-horizon} and $\kappa_z := \mu_z/4$.

\emph{$\vx$-block descent.}
For fixed $\vz_t, \vu_t$ the function $F_t(\vx) + (\rho/2)\norm{\vx - (\vz_t - \vu_t)}^2$
differs from $\mathcal{L}_{\rho,t}(\vx, \vz_t, \vu_t)$ by constants
independent of $\vx$, so the prior-prox inexactness
(Proposition~\ref{prop:x-block}) gives
\[
\mathcal{L}_{\rho,t}(\vx_{t-1}, \vz_t, \vu_t) \;\le\; \mathcal{L}_{\rho,t}(\vx_t, \vz_t, \vu_t) + \varepsilon_t^x.
\tag{S1}
\]

\emph{$\vz$-block descent.}
For fixed $\vx_{t-1}, \vu_t$, let
$Q_t(\vz) = \gamma J_t(\vz; y) + (\rho/2)\norm{\vz - (\vx_{t-1}+\vu_t)}^2$ and
$\vz_t^\star = \argmin_{\vz} Q_t(\vz)$. The planning-prox inexactness
(Proposition~\ref{prop:z-block}) and $\mu_z$-strong convexity give
\[
Q_t(\vz_{t-1}) \;\le\; Q_t(\vz_t^\star) + \varepsilon_t^z,
\qquad
\norm{\vz_{t-1} - \vz_t^\star}^2 \;\le\; \frac{2\varepsilon_t^z}{\mu_z}.
\]
Writing $\vz_t - \vz_t^\star = (\vz_t - \vz_{t-1}) + (\vz_{t-1} - \vz_t^\star)$ and
using $\norm{\va + \vb}^2 \ge \tfrac{1}{2}\norm{\va}^2 - \norm{\vb}^2$,
\[
Q_t(\vz_t) - Q_t(\vz_t^\star) \;\ge\; \tfrac{\mu_z}{4}\norm{\vz_t - \vz_{t-1}}^2 - \varepsilon_t^z.
\]
Combining, $Q_t(\vz_{t-1}) \le Q_t(\vz_t) - \kappa_z \norm{\vz_t - \vz_{t-1}}^2 + 2\varepsilon_t^z$,
so
\[
\mathcal{L}_{\rho,t}(\vx_{t-1}, \vz_{t-1}, \vu_t) \;\le\;
\mathcal{L}_{\rho,t}(\vx_{t-1}, \vz_t, \vu_t) - \kappa_z \norm{\vz_t - \vz_{t-1}}^2 + 2\varepsilon_t^z.
\tag{S2}
\]
Combining (S1) and (S2),
\[
\mathcal{L}_{\rho,t}(\vx_{t-1}, \vz_{t-1}, \vu_t) \;\le\;
\mathcal{L}_{\rho,t}(\vx_t, \vz_t, \vu_t) - \kappa_z \norm{\vz_t - \vz_{t-1}}^2
+ \varepsilon_t^x + 2\varepsilon_t^z.
\tag{S3}
\]

\emph{Dual update.}
The scaled update $\vu_{t-1} = \vu_t + \vr_{t-1}$ raises the Lagrangian by
$\rho\norm{\vr_{t-1}}^2$:
$\mathcal{L}_{\rho,t}(\vx_{t-1}, \vz_{t-1}, \vu_{t-1}) =
\mathcal{L}_{\rho,t}(\vx_{t-1}, \vz_{t-1}, \vu_t) + \rho\norm{\vr_{t-1}}^2$.
The dual-residual domination
$\rho\norm{\vr_{t-1}}^2 \le \theta\norm{\vz_t - \vz_{t-1}}^2 + C_r(\varepsilon_t^z + \varepsilon_{t+1}^z + \omega_t)$
($0 \le \theta < \kappa_z$) then gives
\[
\mathcal{L}_{\rho,t}(\vx_{t-1}, \vz_{t-1}, \vu_{t-1}) \;\le\;
\mathcal{L}_{\rho,t}(\vx_t, \vz_t, \vu_t) - \bar\kappa\, \norm{\vz_t - \vz_{t-1}}^2
+ C_1(\varepsilon_t^x + \varepsilon_t^z + \varepsilon_{t+1}^z + \omega_t),
\]
with $\bar\kappa := \kappa_z - \theta > 0$. Using the domination once more to
control $\norm{\vr_{t-1}}^2$,
\begin{equation*}
\begin{aligned}
\mathcal{L}_{\rho,t}(\vx_{t-1}, \vz_{t-1}, \vu_{t-1})
&\;\le\; \mathcal{L}_{\rho,t}(\vx_t, \vz_t, \vu_t)
- c\,(\norm{\vr_{t-1}}^2 + \norm{\vz_t - \vz_{t-1}}^2) \\
&\quad + C_3(\varepsilon_t^x + \varepsilon_t^z + \varepsilon_{t+1}^z + \omega_t),
\end{aligned}
\tag{S4}
\end{equation*}
for some $c > 0$.

\emph{Drift and telescoping.}
The objective changes from step $t$ to $t-1$. By
\eqref{eq:drift},
$\mathcal{L}_{\rho,t-1}(\vx_{t-1}, \vz_{t-1}, \vu_{t-1})
\le \mathcal{L}_{\rho,t}(\vx_{t-1}, \vz_{t-1}, \vu_{t-1}) + \omega_t$.
Combined with (S4),
\[
\begin{aligned}
c\,(\norm{\vr_{t-1}}^2 + \norm{\vz_t - \vz_{t-1}}^2)
&\;\le\; \mathcal{L}_{\rho,t}(\vx_t, \vz_t, \vu_t) - \mathcal{L}_{\rho,t-1}(\vx_{t-1}, \vz_{t-1}, \vu_{t-1}) \\
&\quad + C_4(\varepsilon_t^x + \varepsilon_t^z + \varepsilon_{t+1}^z + \omega_t).
\end{aligned}
\]
Summing over $t = T, T-1, \dots, 1$ telescopes the Lagrangian terms:
\[
c\sum_{t=1}^T(\norm{\vr_{t-1}}^2 + \norm{\vz_t - \vz_{t-1}}^2)
\;\le\; \mathcal{L}_{\rho,T}(\vx_T, \vz_T, \vu_T) - \mathcal{L}_{\rho,0}(\vx_0, \vz_0, \vu_0)
+ C_5\sum_{t=1}^T(\varepsilon_t^x + \varepsilon_t^z + \omega_t).
\]
replacing $\mathcal{L}_{\rho,0}$ by $\underline{\mathcal{L}}$ (compactness and
lower-boundedness) and taking the minimum over $t$ yields
\eqref{eq:thm3-result}.
\end{proof}

\noindent\emph{Sufficient condition for the domination.}
The approximate $\vz$-stationarity
$\ve_t^z := \gamma\nabla J_t(\vz_{t-1}; y) + \rho(\vz_{t-1} - \vx_{t-1} - \vu_t)$
has $\norm{\ve_t^z}^2 \lesssim \varepsilon_t^z$; with
$\vu_{t-1} = \vu_t + (\vx_{t-1} - \vz_{t-1})$,
$\rho \vu_{t-1} = \gamma\nabla J_t(\vz_{t-1}; y) - \ve_t^z$. Differencing against
the neighbouring step $t$ and using $L_G$-Lipschitz continuity of
$\gamma\nabla J_t$ (with the gradient-drift term inside $\omega_t$),
\[
\rho\,\norm{\vr_{t-1}}^2 \;\lesssim\; \frac{L_G^2}{\rho}\,\norm{\vz_t - \vz_{t-1}}^2
+ \varepsilon_t^z + \varepsilon_{t+1}^z + \omega_t,
\]
so the domination holds with $\theta \approx L_G^2/\rho^2 < \kappa_z$ once
$\rho$ is large relative to $L_G$ and $\mu_z$.

\section{Additional result figures}
\label{app:result-figures}

\begin{figure}[H]
\centering
\includegraphics[width=\linewidth]{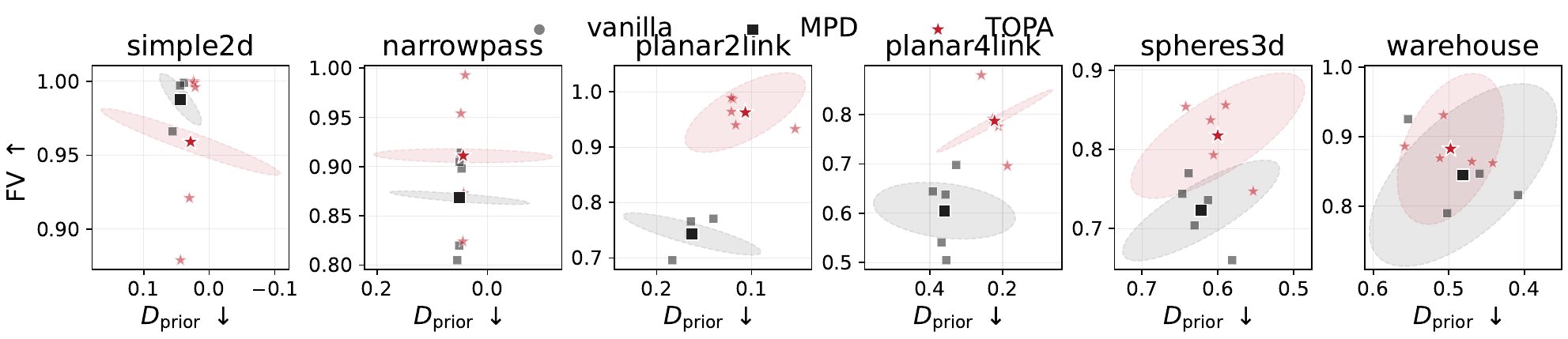}
\caption{Feasibility--fidelity trade-off per task: fraction valid \textbf{FV}
($\uparrow$) against prior deviation $D_{\mathrm{prior}}$ ($\downarrow$; the
$x$-axis is inverted), one panel per task. Markers: vanilla
(\textcolor{gray}{$\bullet$}), MPD ($\blacksquare$), \method{}
({\color{red!70!black}$\bigstar$}); the large marker is the per-method mean and
the dashed ellipse its confidence region. Better is up-and-right, so the
upper-right corner is Pareto-favourable (high feasibility, small drift from the
prior); \method{} occupies it on 4/6 tasks, while on the other two it trades
off---warehouse at higher FV, simple2d at smaller $D_{\mathrm{prior}}$ but
lower FV.}
\label{fig:pareto}
\end{figure}

\begin{figure}[H]
\centering
\includegraphics[width=\linewidth]{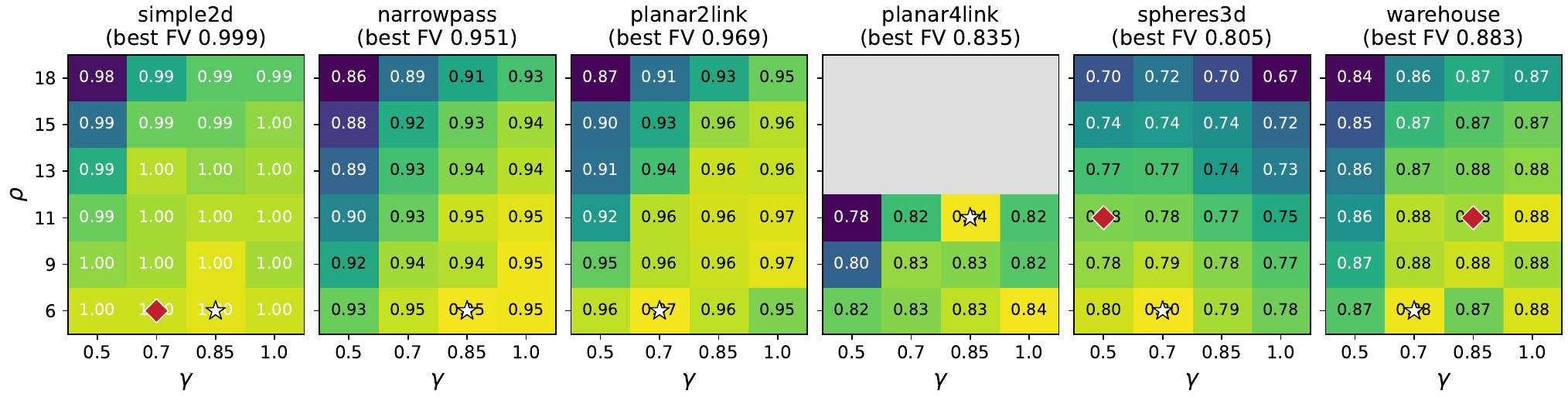}
\caption{Hyperparameter sensitivity per task: fraction valid \textbf{FV}
(printed in each cell and shown as colour; brighter\,=\,higher\,=\,better) over
the ADMM penalty $\rho$ (rows) and the cost-gradient scale $\gamma$
(columns). The white $\bigstar$ marks the best-FV cell and the
{\color{red!70!black}$\blacklozenge$} the configuration we report; gray cells
are not swept. A wide bright plateau around the optimum --- a $2\times2$ block
within 2\,pp of the best FV on 5/6 tasks --- shows the method is
\emph{parameter-robust}.}
\label{fig:heatmap}
\end{figure}

\begin{figure}[p]
\centering
\includegraphics[width=\linewidth,height=0.88\textheight,keepaspectratio]{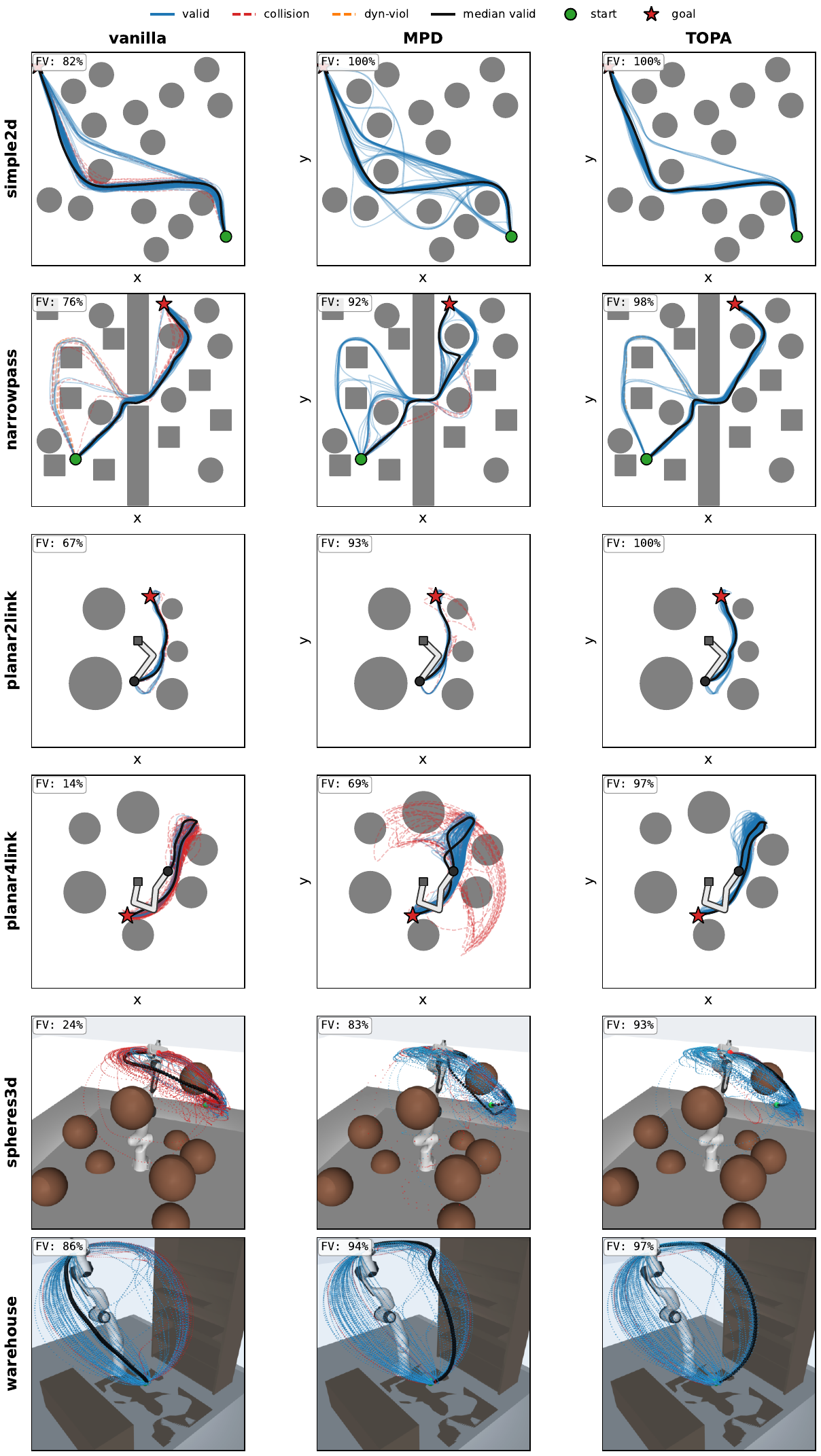}
\caption{Per-sample trajectory distributions, \textcolor{blue}{solid blue} = valid
(collision-free and dynamically feasible), \textcolor{red}{red dashed} =
collision, \textcolor{orange}{orange dashed} = dynamic-limit violation only,
and \textbf{black} = the median valid trajectory; the corner box reports the
valid fraction (FV). More blue and less red is better:
\method{} yields the densest valid bundle and the fewest collisions across
the tasks.}
\label{fig:traj-overlay}
\end{figure}

We collect here the result figures referenced in Sec.~\ref{sec:expts}: the
per-task feasibility--fidelity trade-off (Fig.~\ref{fig:pareto}), the
hyperparameter-sensitivity heatmap (Fig.~\ref{fig:heatmap}), and the per-sample
trajectory distributions (Fig.~\ref{fig:traj-overlay}).

\end{document}